\documentclass[preprint,nopreprintline,12pt]{elsarticle}
\usepackage[T1]{fontenc}
\usepackage[utf8]{inputenc}
\usepackage{lmodern}
\usepackage[a4paper,margin=20mm]{geometry}
\usepackage{amsmath,amssymb,booktabs,array,longtable}
\usepackage{graphicx}
\usepackage{float}
\usepackage{caption}
\newcolumntype{L}[1]{>{\raggedright\arraybackslash}p{\dimexpr#1\linewidth-2\tabcolsep\relax}}
\newcolumntype{R}[1]{>{\raggedleft\arraybackslash}p{\dimexpr#1\linewidth-2\tabcolsep\relax}}
\biboptions{numbers,square,sort&compress}
\journal{}

\pdftrailerid{}
\PassOptionsToPackage{pdfcreator={},pdfproducer={}}{hyperref}

\usepackage[hidelinks]{hyperref}
\makeatletter
\AtBeginDocument{\let\Hy@author\@empty}
\makeatother

\begin{document}
\begin{frontmatter}
\title{\textbf{Robust small-molecule identification from incomplete, degraded, and inconsistent spectra using multimodal mixed-condition training}}

\author[affiliation]{\textbf{Bowen Gao}}
\ead{gaobowend@gmail.com}
\author[affiliation]{\textbf{Lei Zhu}}
\ead{leizhu@mail.sim.ac.cn}
\author[affiliation]{\textbf{Yiying Wang}\texorpdfstring{\corref{corresponding}}{}}
\ead{yywang@mail.sim.ac.cn}
\author[affiliation]{\textbf{Wenjie Yu}}
\ead{casan@mail.sim.ac.cn}
\address[affiliation]{State Key Laboratory of Materials for Integrated Circuits, Shanghai Institute of Microsystem and Information Technology, Chinese Academy of Sciences, 865 Changning Road, Shanghai 200050, China}
\cortext[corresponding]{Corresponding author.}

\begin{abstract}
Reliable small-molecule identification often requires complementary evidence from multiple spectroscopic measurements, and recent advances in artificial intelligence have provided new means to integrate and analyze such information. In practice, however, spectra may be unavailable, degraded by measurement-related variations, or even incorrectly associated with a sample, thereby hindering accurate molecular identification. Herein, we propose a multimodal mixed-condition training strategy that accommodates missing, degraded, and mismatched measurements for small-molecule structure identification. The strategy incorporates chemical and spectroscopic knowledge through predefined missing-input configurations, modality-specific spectral perturbations, and chemically informed spectrum replacements. Models were trained on 635,441 samples comprising mass spectrometry (MS), infrared (IR), and nuclear magnetic resonance (NMR) simulated spectra from the Multimodal Spectroscopic Dataset (MSSD). They were then systematically evaluated on 79,462 held-out samples across 30 views designed to represent variations in spectra. A controlled comparison of complete-input and mixed-condition training under concatenation and mixture-of-experts (MoE) fusion showed that the training strategy was the principal source of improvement. For MoE, mixed-condition training increased the mean reciprocal rank (MRR) by 6.08\% (from 0.9203 to 0.9763) and the top-1 molecular identification rate by 7.67\% (from 89.50\% to 96.36\%). Notably, under single-modality inputs, IR MRR increased 2.15-fold (from 0.4337 to 0.9307), while MS MRR increased 2.31-fold (from 0.3711 to 0.8575). With the proposed strategy, complete-input performance remained high, while sample-level mismatch detection also improved. Together, these results highlight the potential of multimodal mixed-condition training for practical molecular identification by explicitly addressing incomplete, degraded, and mismatched measurements encountered in real-world analysis.
\end{abstract}

\begin{keyword}
Chemometrics \sep Multimodal spectroscopy \sep Small-molecule identification \sep Molecular structure reranking \sep Deep learning \sep Chemistry-informed learning
\end{keyword}
\end{frontmatter}

\section{Introduction}
\label{sec:introduction}

Small-molecule structure identification relies on spectroscopic measurements and their interpretation. Structural confirmation commonly requires several complementary measurements, since each probes a different aspect of the molecule. For example, mass spectrometry (MS) can provide fragmentation patterns that inform substructures and connectivity; infrared (IR) spectroscopy reflects molecular vibrations associated with functional groups and bonding environments; nuclear magnetic resonance (NMR) spectroscopy provides chemical shifts and coupling information associated with the local environments of the observed nuclei. Structure elucidation therefore requires joint analysis of these complementary spectroscopic observations~\cite{priessner2026mmst}. However, joint interpretation of these spectra has traditionally relied on accumulated expertise and time-consuming manual analysis~\cite{mirza2026secs,gao2021graphyne,gao2021mlsynthesis}.

With recent advances in artificial intelligence (AI) and the development of multimodal spectroscopic datasets, AI-accelerated structure elucidation based on complementary spectroscopic evidence has become increasingly feasible~\cite{zou2023qm9s,alberts2024mssd,tan2025clams,priessner2026mmst,mirza2026secs,huang2025unipoly}. For example, Zou et al. developed QM9S, which provides calculated infrared (IR), Raman, and ultraviolet--visible (UV--Vis) spectra for approximately 0.13 million small organic molecules~\cite{zou2023qm9s}. Using a QM9S-derived dataset, Tan subsequently developed CLAMS, a Transformer-based generative model that combined IR, UV--Vis, and \textsuperscript{1}H~NMR spectra and achieved Top-1 and Top-15 structure-generation accuracies of 45.2\% and 83.1\%, respectively, on held-out simulated data~\cite{tan2025clams}. Alberts et al. at IBM Research developed the Multimodal Spectroscopic Dataset (MSSD), which pairs approximately 0.79 million molecular structures with simulated IR, NMR, and MS spectra, a scale more readily achievable through simulation than through fully paired experimental measurements~\cite{alberts2024mssd}. The accompanying MSSD benchmark employed an encoder--decoder Transformer to combine \textsuperscript{1}H~NMR and \textsuperscript{13}C~NMR spectra with a molecular-formula prior, achieving Top-1 and Top-10 structure-prediction accuracies of 73.38\% and 89.98\%, respectively, on simulated data. In a retrieval setting, Mirza et al. developed SECS, which aligned spectral and molecular representations through contrastive learning and recovered the correct structure at rank 1 for 98.4\% of 1,000 MSSD-derived test molecules by combining IR, \textsuperscript{1}H~NMR, \textsuperscript{13}C~NMR, and heteronuclear single quantum coherence (HSQC) spectra~\cite{mirza2026secs}. Beyond these dataset-centered studies, Priessner et al. developed the MultiModalSpectralTransformer (MMST), which was trained on a larger collection of simulated spectra for 4 million ZINC compounds and achieved Top-1 and Top-3 accuracies of 72\% and 80\%, respectively, on a 4,000-compound simulated test subset after HSQC matching and candidate ranking~\cite{priessner2026mmst}. Collectively, these approaches integrate complementary spectral information for molecular structure elucidation. In practice, however, limited experimental resources and sample constraints may prevent the acquisition of a complete set of spectra for a given sample. Variations in sample preparation and instrument settings can introduce noise, and accidental sample mismatches may also occur~\cite{burns2021nmrerrors,lu2017metabolite,fu2025samplemixups}. There is therefore an urgent need to develop targeted training strategies that improve the robustness of multimodal models to incomplete, noisy, or mismatched spectral inputs, systematically evaluate their effects on predictive performance, and quantify the resulting robustness gains~\cite{Beese2024factor}.

Herein, we introduce a multimodal mixed-condition training strategy (denoted as mixed training) for robust small-molecule identification within a candidate structure reranking framework. Chemical and spectroscopic domain knowledge guides the design of the training conditions, which combine predefined missing-input configurations, modality-specific spectral perturbations, and chemically informed spectrum replacements to represent variations in spectral availability, quality, and molecular-source consistency. To distinguish the contribution of the training strategy from that of the fusion architecture, we conducted a controlled two-by-two factorial comparison by crossing complete-input training (denoted as clean training) and mixed training with vanilla concatenation (Concat) and mixture-of-experts (MoE) fusion. The results show that mixed training is the principal source of robustness gains across both architectures, particularly under incomplete, degraded, or mismatched inputs, while maintaining high performance with complete, unaltered inputs. By explicitly training models to accommodate spectral variations, the proposed strategy offers a systematic approach to improving the reliability of multimodal molecular identification under practical analytical conditions.

\section{Methods}
\label{sec:methods}

\subsection{Candidate structure reranking task}
\label{sec:task}

\begin{figure}[H]
\centering
\includegraphics[width=\linewidth,height=0.68\textheight,keepaspectratio]{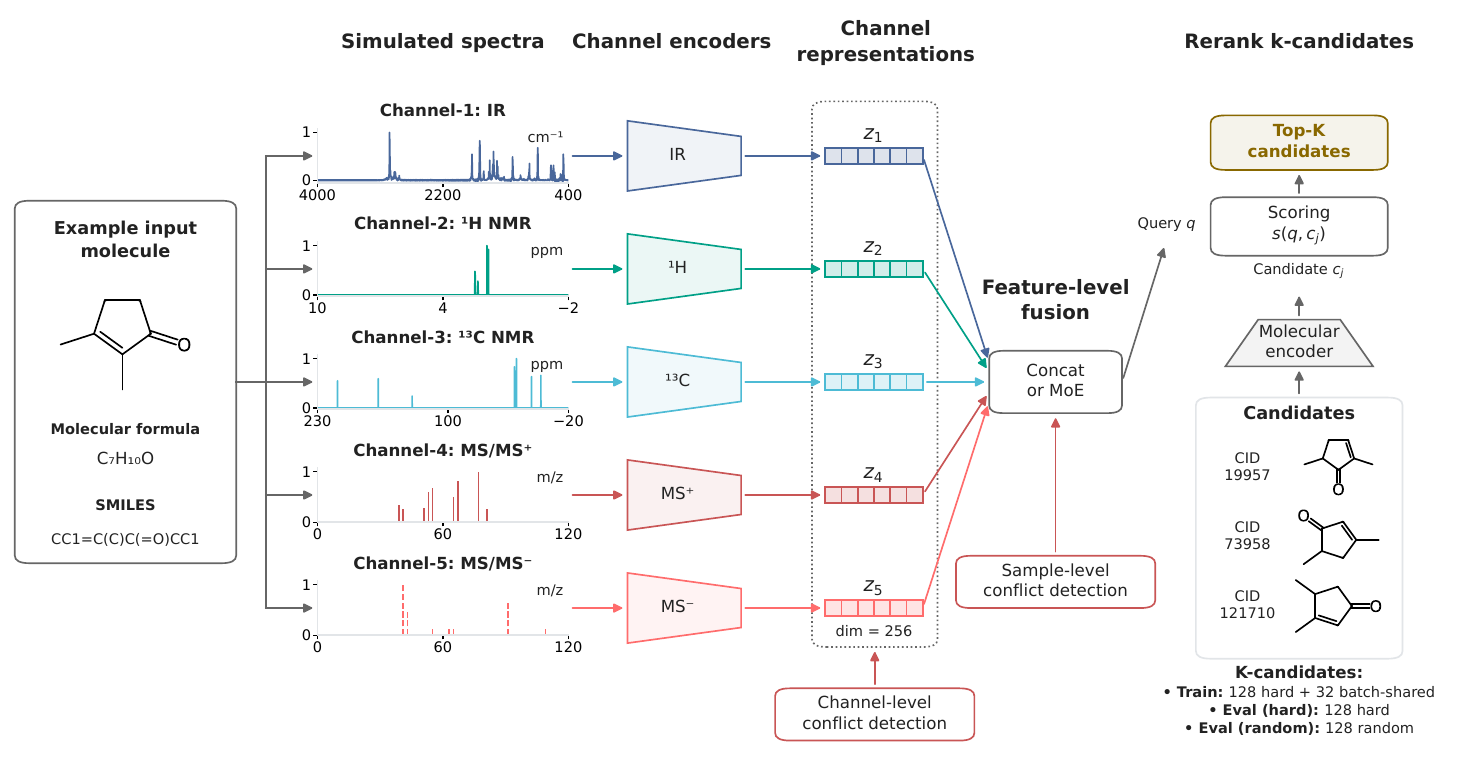}
\caption{Candidate structure reranking from multimodal spectra. Channel- and sample-level conflict detection is discussed in Section~\ref{sec:training-effects}. Five channel encoders and feature-level fusion produce a query representation for scoring encoded candidate structures. The hard candidate lists comprise the correct structure with up to 127 structurally similar negative candidates; 32 batch-shared random negatives are added only during training. Evaluation with random candidates uses the fixed MoE-mixed checkpoint and 127 negatives sampled globally at random. Fusion architectures are detailed in Figs.~S1--S2.}
\label{fig:1}
\end{figure}

The task was to rank a finite set of candidate molecular structures using a query's available spectra (Fig.~\ref{fig:1}). For each query, the set contained the correct structure and a collection of alternative structures. Spectral encoders and a fusion module produced a query representation, while a molecular encoder represented each candidate (Tables~S1--S2). A compatibility score then determined the candidate order. It is worth noting that formula features were computed independently for each candidate structure as part of its molecular representation. The correct molecular formula of the query was not provided to the model as prior information. In summary, the task assessed discrimination within the supplied candidate set, conditional on inclusion of the correct structure.

\subsection{Spectroscopic data and candidate sets}
\label{sec:data}

\subsubsection{Data sources and splits}

MSSD version~3, comprising 794,386 aligned records of simulated multimodal spectra, was used in this study~\cite{alberts2024mssd,wang2004gaff,thompson2022lammps,cobas2020nmr,wang2021cfmid4}. Four spectroscopic modalities were represented across five input channels: IR, \textsuperscript{1}H~NMR, \textsuperscript{13}C~NMR, and liquid chromatography–electrospray ionization tandem mass spectrometry (LC-ESI-MS/MS) in positive- and negative-ion modes (ESI\textsuperscript{+} and ESI\textsuperscript{-}). The two mass-spectral channels were predicted using Competitive Fragmentation Modeling for Metabolite Identification (CFM-ID) at a collision energy of 40~eV, with protonated and deprotonated precursor ions assigned as $\mathrm{[M+H]}^{+}$ and $\mathrm{[M-H]}^{-}$, respectively~\cite{wang2021cfmid4}.

\begin{figure}[H]
\centering
\includegraphics[width=100mm]{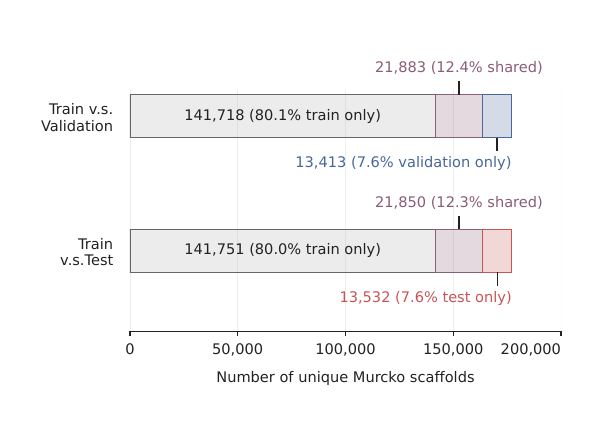}
\caption{Murcko scaffold analysis identifies shared and distinct molecular cores in training--validation and training--test comparisons.}
\label{fig:2}
\end{figure}

Following preprocessing and eligibility assessment across all five channels, 794,343 samples were retained and partitioned for the study (Table~\ref{tab:1}). Records sharing the same molecular connectivity, including stereochemical variants, were assigned to the same dataset split. International Chemical Identifier Keys (InChIKey) were generated from molecular structures represented by canonical SMILES, and molecules were grouped by the first InChIKey block (connectivity block)~\cite{heller2015inchi}. The resulting groups were assigned to the training, validation, and test sets to achieve an approximate 80:10:10 split. The validation and test sets contained 13,413 and 13,532 unique nonempty Murcko scaffolds absent from training, respectively, demonstrating distinct scaffold coverage despite remaining shared scaffolds (Fig.~\ref{fig:2})~\cite{bemis1996frameworks}.

\begin{table}[H]
\centering
\small
\caption{Dataset partition sizes after multimodal alignment. Counts refer to eligible samples with all five input channels within each partition.}\label{tab:1}
\setlength{\tabcolsep}{4pt}
\renewcommand{\arraystretch}{1.16}
\begin{tabular}{@{}L{0.50}R{0.50}@{}}
\toprule
Split & Samples \\
\midrule
Train & 635,441 \\
Validation & 79,440 \\
Test & 79,462 \\
\bottomrule
\end{tabular}
\end{table}

\subsubsection{Candidate set construction}

Negative candidates were drawn from an external pool of approximately 14.5 million deduplicated molecular structures constructed from PubChem~\cite{kim2025pubchem}. Structures were standardized with RDKit version~2026.3.2 to canonical isomeric SMILES strings, preserving stereochemical information present in the source structures. For the evaluation, a neutral exact-mass index window was used to obtain alternatives near the mass of the correct structure, with chemically related hard negatives prioritized within that window (Fig.~\ref{fig:1}, $K$-candidates). Candidate identities were held fixed across spectral perturbations. Further details are provided in Section~S2 of the supplementary material. 

\subsection{Input conditions and spectral perturbations}
\label{sec:conditions}

\begin{figure}[H]
\centering
\includegraphics[width=\linewidth,height=0.68\textheight,keepaspectratio]{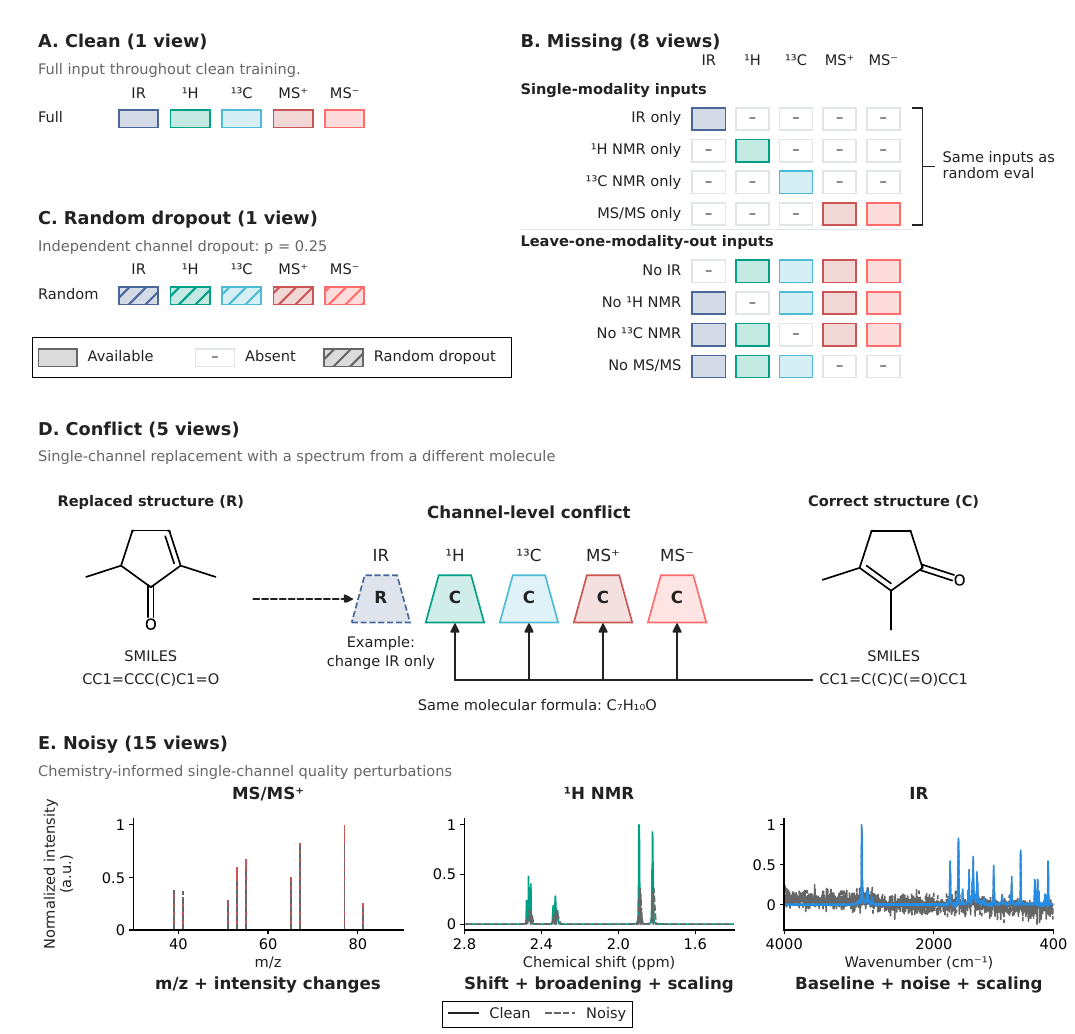}
\caption{Domain-informed input perturbations covering spectral availability, quality, and molecular-source consistency. (A) Clean input retains all five unaltered channels. (B) Missing-input conditions display eight fixed masks, and the random eval label marks the four single-modality masks used in the candidate-condition comparison (details in Section~3.4). (C) Random dropout independently removes each of the five channels with a probability of~0.25. (D) The structure pair illustrates replacement of the IR source while retaining the other four channels from the correct structure. (E) Paired spectra illustrate separate single-channel, high-severity perturbations. Registered view families and perturbation parameters are defined in Tables~S3--S4.}
\label{fig:3}
\end{figure}

A set of 30 predefined input conditions (hereafter termed views) was used to cover variations in spectral availability, signal quality, and cross-modal consistency encountered in practical spectroscopic measurements (Fig.~\ref{fig:3}). The 30 views comprised 1 clean (complete input), 8 missing-input, one random-dropout, 15 noisy, and 5 conflict views. The implementation and full parameter set are described in Section~S1.1 and Tables~S3--S4. Briefly, clean view (Fig.~\ref{fig:3}A) included all five unaltered input channels. The eight missing-input views (Fig.~\ref{fig:3}B) represented four single-modality and four leave-one-modality-out configurations, without additional numerical perturbation or source replacement. A binary availability mask determined which channel embeddings were included during fusion. In the separate random-dropout view (Fig.~\ref{fig:3}C), each of the five channels was independently removed with a probability of~0.25.

Conflict views replaced one channel with the corresponding spectrum from another MSSD molecule in the same partition, labeled as replaced structure (R). Replacement-source selection emphasized shared molecular formula, scaffold, structural similarity, or functional-group composition, supplemented by a smaller random-replacement component. This chemically informed construction introduced incorrect-source evidence that could remain plausible alongside spectra from the correct structure (C). The other four channels continued to originate from the correct structure (Fig.~\ref{fig:3}D). Availability masks, perturbed inputs, and replacement-source assignments were generated and fixed before model evaluation. Consequently, the four models encountered the same fixed views for every test sample.

Noisy views (Fig.~\ref{fig:3}E and Fig.~S3) altered one channel at a time at low, medium, or high severity while retaining the other four channels. Perturbations were matched to the representation and signal characteristics of each modality. MS/MS perturbations combined mass-to-charge ratio ($m/z$) jitter, intensity variation, stochastic peak removal, and exclusion of low-intensity peaks~\cite{loffler2025massaccuracy,huber2021ms2deepscore,dallavalle2025noisefiltering}. NMR perturbations comprised chemical-shift displacement, line broadening, intensity scaling, and signal or annotation dropout~\cite{savorani2010icoshift,li2021deeppicker,li2026deepphaser}. IR perturbations combined baseline variation, additive Gaussian noise, intensity scaling, and removal of individual sampled points~\cite{sy2024voccertifire,conlin1998augmentation,blazhko2021emsa,Wohlers2023augmenting}. The numerical operations and severity levels were fixed in advance, allowing a consistent comparison across models. These operations provided controlled tests of changes in spectral quality, with full parameter definitions in Table~S4.

\subsection{Model architectures and training strategies}
\label{sec:models}

All four models shared the molecular and spectral encoders summarized in Table~S1. Candidate structures were represented by ChemBERTa embeddings and candidate-derived chemical features~\cite{ahmad2022chemberta2}. DreaMS encoded both MS/MS channels~\cite{bushuiev2026dreams}. Local Transformer encoders were used to process IR and NMR spectra, due to implementation compatibility and pretrained-weight availability~\cite{huang2025act,han2026nmrvit}. Naive concatenation combined the ordered channel representations with missing inputs zeroed, and MoE adaptively pooled the available representations through selected experts~\cite{shazeer2017moe}. The fused representation was projected into the candidate embedding space for compatibility scoring. Further details are provided in Section~S1 and Figs.~S1--S2.

The factorial comparison crossed Concat or MoE fusion with clean or mixed-view training. Clean training used complete, unperturbed inputs throughout. Mixed training first sampled complete and missing inputs, then introduced noisy and conflict views while retaining complete and sparse inputs. Both strategies optimized candidate cross-entropy with auxiliary chemical-feature and evidence-consistency objectives. MoE also used load balancing~\cite{fedus2022switch}. Each combination had one fitted model, with matched total training steps and global batch size. One checkpoint per run was selected by validation MRR on the clean view and evaluated on the same held-out views and hard candidates. Detailed computational settings are listed in Table~S2.

\subsection{Evaluation metrics and statistical analysis}
\label{sec:metrics}

Mean reciprocal rank (MRR) was the primary ranking metric, supplemented by recall at ranks 1, 5, and 10. Expected calibration error (ECE) measured confidence calibration. Areas under the receiver operating characteristic and precision--recall curves (AUROC and AUPRC) measured sample-level conflict detection~\cite{guo2017calibration}. Channel localization assessed identification of the replaced channel. Ranking excluded padded candidates and assigned the best position within exact score ties. Alternative tie conventions were examined separately (Section~S3.3).

Overall performance was summarized by assigning equal weight to all 30 views. Training, fusion, and interaction effects used paired comparisons. Percentile 95\% confidence intervals came from 1,000 bootstrap resamples of test sample identities, preserving pairing across models and views~\cite{efron1986bootstrap}. They quantify test-sample uncertainty for the fixed checkpoints. Metric definitions and statistical procedures are given in Section~S3.1, and interval estimates are reported in Table~S5.

\section{Results and discussion}
\label{sec:results}

\subsection{Overall performance}

\begin{figure}[H]
\centering
\includegraphics[width=\linewidth,height=0.68\textheight,keepaspectratio]{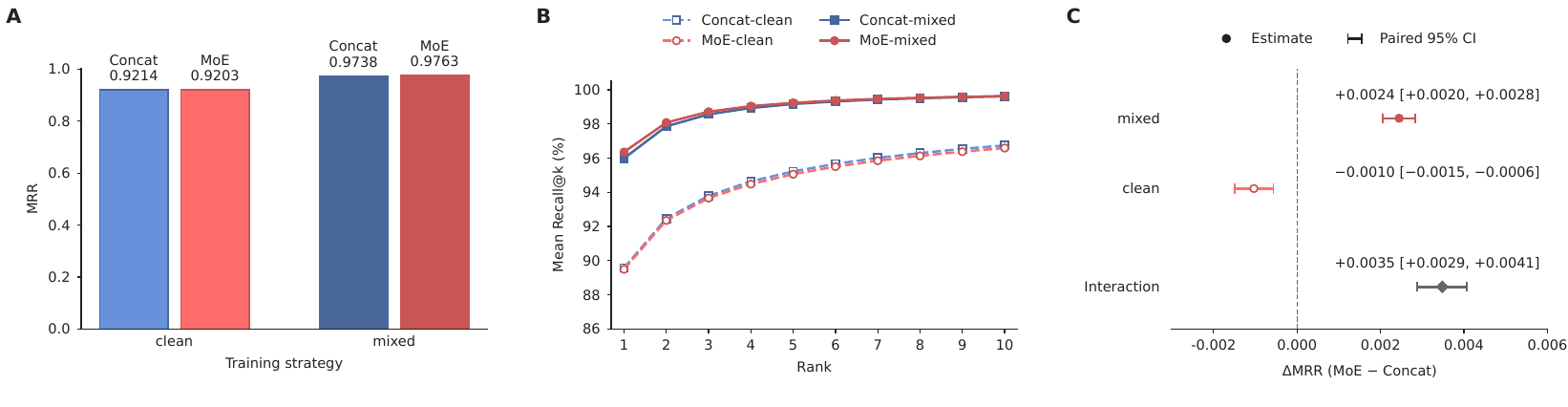}
\caption{Overall reranking performance with hard candidates ($n=79{,}462$). (A) MRR and (B) Recall@$k$, averaged equally over 30 views. (C) Fusion effects (MoE $-$ Concat) and their interaction (mixed $-$ clean). Error bars denote paired-bootstrap 95\% confidence intervals.}
\label{fig:4}
\end{figure}

On the whole, the mixed training protocol provided the largest overall improvement in candidate reranking, and MoE supplied an additional gain within that training strategy (Fig.~\ref{fig:4}A and Table~\ref{tab:2}). The MoE-mixed model achieved a 30-view mean MRR of 0.9763 and recall at rank 1 (R@1) of 96.36\%, compared with 0.9203 and 89.50\%, respectively, for MoE-clean. These changes corresponded to a 6.08\% relative increase in MRR and a 6.86~pp increase in R@1, or a 7.67\% relative increase in first-rank recovery. Concat showed the same main pattern, reaching an MRR of 0.9738 and R@1 of 95.99\% with mixed training. The common direction across architectures supports the combined benefit of the complete mixed training protocol.

The improvement also extended from first-rank placement to recovery within a short candidate list (Fig.~\ref{fig:4}B). For MoE-mixed, average R@5 and R@10 were 99.24\% and 99.63\%, respectively, indicating that the correct structure was usually recovered within the first ten candidates under the predefined views. The corresponding MoE-clean values were 95.06\% and 96.59\%. Taken together, the MRR and Recall@$k$ comparisons support improved candidate prioritization across several rank cutoffs. The paired MRR training effect was +0.0559 for MoE and +0.0525 for Concat (Table~S5). By comparison, the fusion effect was smaller and depended on the training strategy: MoE modestly outperformed Concat under mixed training but not under clean training, yielding a positive interaction (Fig.~\ref{fig:4}C). Section~3.2 examines the training gains across individual input views, whereas Section~3.3 examines this fusion interaction in detail.

\begin{table}[tbp]
\centering
\footnotesize
\caption{Ranking, calibration, and conflict-detection performance with hard candidates.}\label{tab:2}
\setlength{\tabcolsep}{4pt}
\renewcommand{\arraystretch}{1.16}
\begin{tabular}{@{}L{0.14}L{0.08}R{0.0975}R{0.0975}R{0.0975}R{0.0975}R{0.0975}R{0.0975}R{0.0975}R{0.0975}@{}}
\toprule
Training & Fusion & MRR & R@1 (\%) & R@5 (\%) & R@10 (\%) & ECE (\%) & Clean-view MRR & Conflict AUROC & Conflict AUPRC \\
\midrule
mixed & MoE & \textbf{0.9763} & \textbf{96.36} & \textbf{99.24} & \textbf{99.63} & \textbf{1.57} & \textbf{0.9867} & \textbf{0.8148} & \textbf{0.8368} \\
 & Concat & 0.9738 & 95.99 & 99.17 & 99.62 & 1.66 & 0.9850 & 0.8546 & 0.8801 \\
clean & MoE & 0.9203 & 89.50 & 95.06 & 96.59 & 8.13 & 0.9850 & 0.4961 & 0.4984 \\
 & Concat & 0.9214 & 89.57 & 95.22 & 96.75 & 7.74 & 0.9842 & 0.5002 & 0.5001 \\
\bottomrule
\end{tabular}
\par\medskip
\begin{minipage}{\linewidth}\footnotesize
MRR, recall, and ECE are unweighted 30-view means; clean-view MRR uses complete input. AUROC and AUPRC are averaged over five replacement views, each with 79,462 matched conflict--clean pairs. Higher values indicate better performance, except for ECE. Bold denotes MoE-mixed. Paired effects: Table~S5.
\end{minipage}
\end{table}

\subsection{Effects of training strategy}
\label{sec:training-effects}

\begin{figure}[H]
\centering
\includegraphics[width=0.95\linewidth,height=0.68\textheight,keepaspectratio]{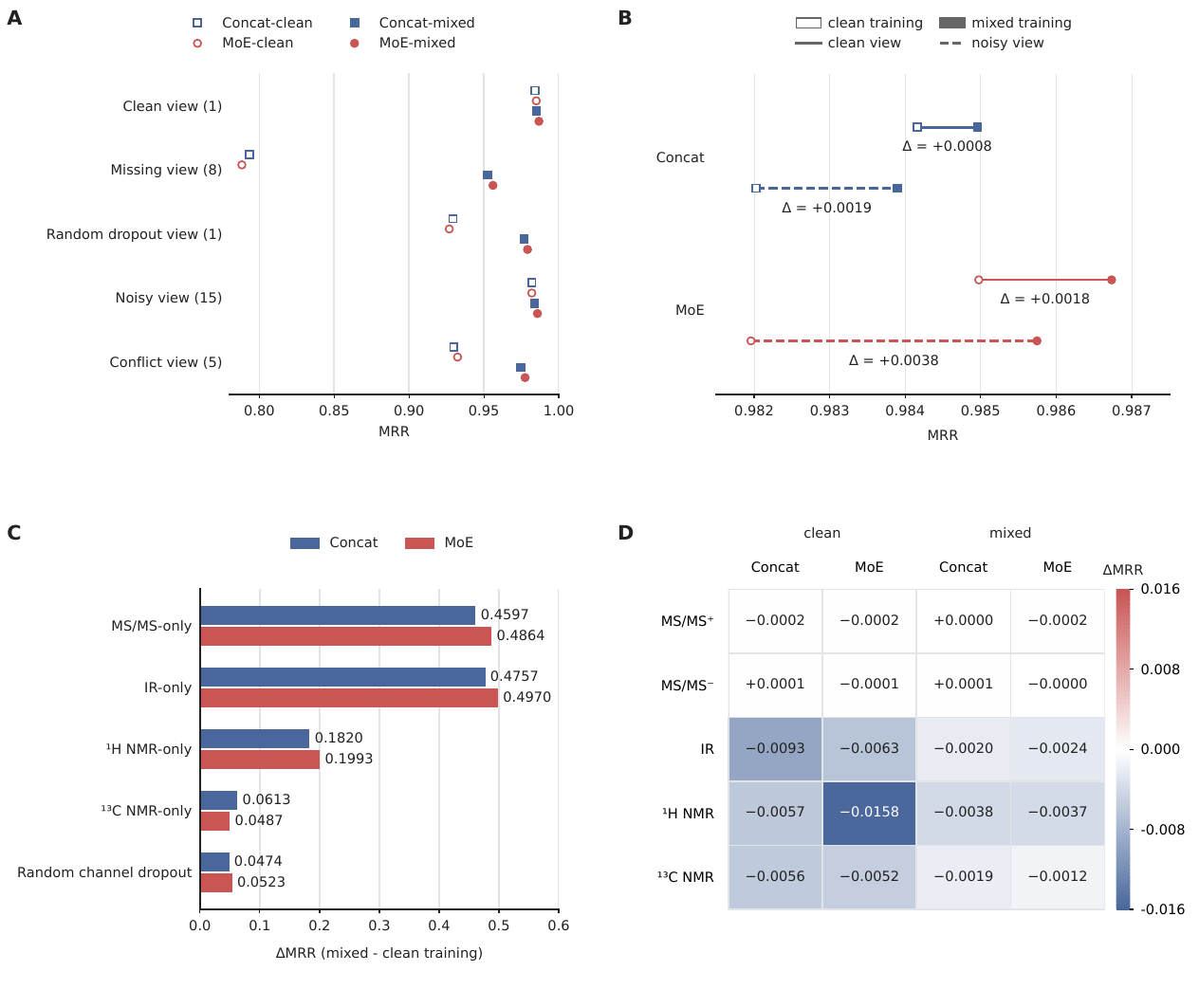}
\caption{Training effects on MRR with hard candidates. (A) View-family means. (B) Clean-view (solid) and noisy-view means (dashed); open/filled markers denote clean/mixed training. (C) Training gains for single-modality inputs and random dropout. (D) Highest-severity single-channel MRR minus each model's clean-view MRR, with other channels unchanged; negative values indicate degradation.}
\label{fig:5}
\end{figure}

Overall, the training effects followed a clear pattern: the largest gains were observed when the available evidence was sparse or incomplete, while high complete-input performance was retained (Fig.~\ref{fig:5}A--C). Specifically, (1) the MoE missing-family mean MRR increased from 0.7883 to 0.9560, corresponding to a 1.21-fold improvement; (2) random-dropout MRR rose from 0.9269 to 0.9791, corresponding to a 5.63\% relative improvement; (3) under complete input, the improvements were marginal, with MoE increasing from 0.9850 to 0.9867 and Concat from 0.9842 to 0.9850 (Fig.~\ref{fig:5}B). The clean family contains only the clean view, which provides the complete-input reference for these comparisons. In summary, the results shown in Fig.~\ref{fig:5}A--C indicate that the much larger gains under reduced input were obtained while high complete-input performance was retained, with no observed decrease for either fitted architecture.

To examine limited evidence more directly, we next compared single-modality inputs (Fig.~\ref{fig:5}C and Fig.~\ref{fig:6}A). First, MoE IR-only MRR increased from 0.4337 to 0.9307, reaching 2.15 times the clean-training value. Second, MS/MS-only MRR increased from 0.3711 to 0.8575, or 2.31 times its baseline. Third, the \textsuperscript{1}H~NMR-only score rose from 0.7633 to 0.9626 (Table~S6), whereas the MoE gain for \textsuperscript{13}C~NMR-only input was smaller (+0.0487 MRR). Concat exhibited corresponding improvements, including IR-only MRR from 0.4374 to 0.9131 and MS/MS-only MRR from 0.4034 to 0.8631. Because these evaluations used the same fitted multimodal models with reduced input, the results suggest that the mixed protocol helped preserve candidate discrimination when other measurements were unavailable, consistent with reduced dependence on a fixed combination of spectra.

Next, random channel dropout extended this observation beyond the fixed input masks, with a MoE training gain of +0.0523 MRR (Fig.~\ref{fig:5}C). The gain depended on the evidence that remained. When both MS/MS channels were removed, clean-training performance was already high, and MRR increased from 0.9834 to 0.9862. By comparison, removing \textsuperscript{13}C~NMR yielded a larger improvement, from 0.8977 to 0.9814 (Table~S6). Accordingly, the family mean in Fig.~\ref{fig:5}A should be read alongside the selected contrasts in Fig.~\ref{fig:5}C and the complete view-by-model comparison in Table~S6.

The results for spectral-quality degradation showed a different but related pattern. All four models retained high MRR when one channel was degraded and the remaining channels were available. For MoE, the noisy-family mean increased from 0.9820 with clean training to 0.9857 with mixed training (Fig.~\ref{fig:5}A and B). At the highest perturbation level, \textsuperscript{1}H~NMR degradation produced MRR values of 0.9692 and 0.9830 for the two strategies, respectively. The corresponding IR values were 0.9787 and 0.9843. These absolute scores reflect both the clean-view baseline and the response to degradation. To isolate the latter, Fig.~\ref{fig:5}D subtracts each model's own clean-view MRR. For MoE, the \textsuperscript{1}H~NMR decrease narrowed from 0.0158 to 0.0037, and the IR decrease from 0.0063 to 0.0024. Smaller IR and NMR decreases were also observed for Concat, whereas MS/MS changes were close to zero under both strategies. Thus, the results in Fig.~\ref{fig:5}B and D jointly support higher noisy-input performance, with reduced degradation for IR and NMR. The complete curves in Fig.~S3 and Table~S6 include some nonmonotonic responses and describe the specified perturbations with complementary channels present.

The near-zero MS/MS changes and larger IR and NMR decreases in Fig.~\ref{fig:5}D motivate a possible spectroscopic interpretation, which can be separated into two points: (1) In ESI-MS/MS, soft ionization produces precursor ions~\cite{cox2014spin}, while subsequent collision-induced dissociation generates fragment populations that depend on molecular structure, precursor-ion type, collision energy, and instrument conditions~\cite{stein2012libraries}. CFM-ID represents this fragmentation process probabilistically by assigning probabilities to competing fragmentation pathways~\cite{wang2021cfmid4}. Consequently, the simulated MS/MS spectra already encode built-in variability associated with alternative fragmentation outcomes, which may reduce the model's dependence on highly specific peak patterns and make its learned representation less sensitive to additional numerical perturbations; (2) IR bands reflect molecular vibrations and bonding environments, whereas one-dimensional \textsuperscript{1}H and \textsuperscript{13}C NMR chemical shifts primarily constrain local nuclear environments~\cite{huang2025act,cobas2020nmr}. Reliance on these relatively localized spectral features may therefore make molecular ranking more sensitive to perturbations in chemical-shift positions or spectral profiles. Under this interpretation, the near-zero MS/MS changes may reflect both the intrinsic fragmentation variability represented by CFM-ID and the preservation of informative fragment constraints under the applied perturbations, with complementary channels providing additional compensation. Here, NMR perturbations included chemical-shift displacement, whereas IR perturbations altered baseline, noise, intensity, and sampled points without explicitly shifting band positions. Given these modality-specific settings, this interpretation concerns simulated spectra with complementary channels retained rather than perturbations matched for equivalent information loss.

Finally, mixed training improved ranking with incorrect-source evidence, where an available spectrum belonged to another molecule rather than being numerically degraded. The conflict-family MRR for MoE rose from 0.9324 to 0.9775, and Concat improved from 0.9298 to 0.9747 (Fig.~\ref{fig:5}A). The greatest MoE gain among the five replacement views occurred for \textsuperscript{13}C~NMR, where MRR increased from 0.8039 to 0.9710. Two benefits were observed: (1) prioritizing the correct candidate despite a mismatch; (2) detecting that the input set contained one. For the second benefit, MoE AUROC increased from 0.4961 to 0.8148, while Concat increased from 0.5002 to 0.8546 (Table~\ref{tab:2}). The corresponding AUPRC values for MoE-mixed and Concat-mixed were 0.8368 and 0.8801. Mean channel-localization accuracy under mixed training was 20.38\% for MoE and 20.78\% for Concat, close to the 20.00\% random-choice level across five channels (Table~S7). Taken together, these results support improved candidate prioritization and sample-level mismatch screening.

In addition, calibration improved alongside ranking. Mean ECE decreased from 8.13\% to 1.57\% for MoE and from 7.74\% to 1.66\% for Concat (Table~\ref{tab:2}). The largest calibration errors for the clean-trained models occurred with sparse evidence, including IR-only and MS/MS-only inputs, and were substantially reduced after mixed training (Table~S6). Improved ordering and lower calibration error address complementary aspects of candidate prioritization: which structures are ranked highest and how well the reported confidence agrees with correctness. Taken together, the view-level results support a training mixture that covers anticipated changes in evidence availability, quality, and consistency while retaining complete-input examples.

\subsection{Fusion architecture effects and interactions}

Having established the dominant effect of training strategy, we next examined whether the fusion architecture contributed an additional gain (Fig.~\ref{fig:4}C). Under mixed training, MoE exceeded Concat by 0.0024 MRR, whereas under clean training its difference from Concat was $-0.0010$. The resulting interaction was +0.0035, indicating that the fusion comparison changed in favor of MoE under the mixed protocol (Table~S5). Consistent with the aggregate comparison in Fig.~\ref{fig:4}A and B, MoE achieved higher MRR than Concat in 29 of the 30 predefined views under mixed training. The largest local gain was +0.0176 for IR-only input. MS/MS-only input was the exception, with a difference of $-0.0056$ (Table~S6).

The relative magnitudes of these effects establish a clear design priority. First, changing the training strategy improved MRR by 0.0525--0.0559 across the two architectures. Second, changing fusion within mixed training added 0.0024. Concat already combined learned spectral representations through a trainable projection and could accommodate the missing-input patterns encountered during training. MoE supplied an additional mechanism for conditioning fusion on the available representations and mask. Therefore, its additional ranking gain is consistent with a benefit from adaptive fusion once diverse input conditions have been included in training. The comparison prioritizes the complete mixed protocol, with fusion refinement providing an additional improvement. Within mixed training, MoE achieved higher ranking scores, whereas Concat achieved higher conflict-detection scores.

The role of domain knowledge lies in the definition of the training protocol. Specifically, (1) general strategies such as modality dropout and data augmentation were implemented using modality-specific signal changes and replacement-source selection based on molecular relationships~\cite{neverova2016moddrop,conlin1998augmentation,Mishra2021synergistic}; (2) a defined sampling schedule, shared hard candidates, and fixed evaluation views made these choices reproducible and allowed their combined benefit to be quantified. The two-by-two factorial comparison therefore supports the mixed protocol as a complete training strategy.

Finally, routing diagnostics were consistent with broader expert use on average after mixed training. The 30-view mean normalized expert-selection entropy increased from 0.7510 to 0.8520, and the mean maximum expert-selection share decreased from 0.4155 to 0.3113 (Table~S8). Missing-family entropy decreased from 0.7147 to 0.6817 alongside improved ranking. These patterns are consistent with condition-dependent routing. Expert identifiers are local to each fitted model, and expert-selection frequencies are distinct from spectral-channel weights. Section~S3.2 further distinguishes these diagnostics from the load-balancing objective, which also depends on router probabilities. The directions of the training, fusion, and interaction effects were preserved under alternative tie conventions (Table~S9).

\subsection{Modality contributions and candidate set composition}

\begin{figure}[H]
\centering
\includegraphics[width=\linewidth,height=0.68\textheight,keepaspectratio]{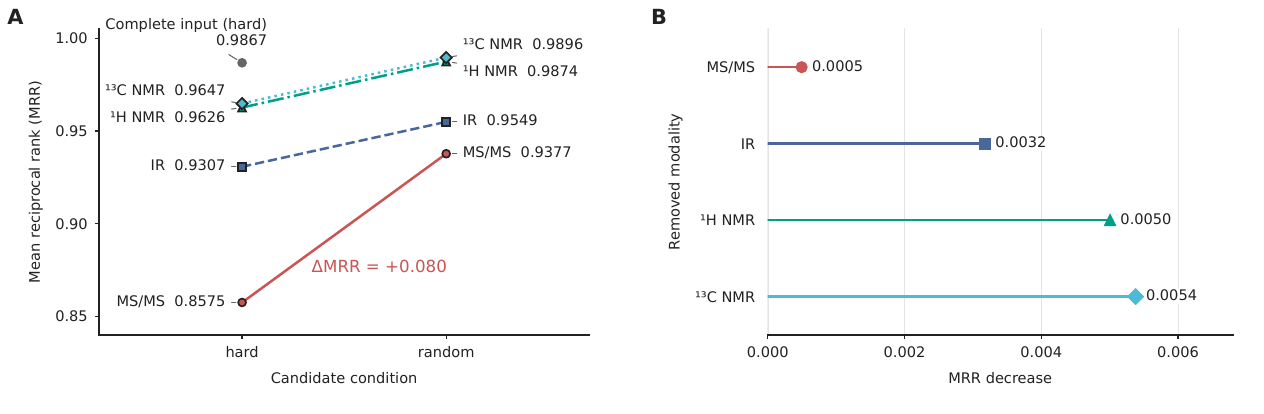}
\caption{Modality contributions for the fixed MoE-mixed checkpoint. (A) Single-modality MRR with hard and random candidates; the gray reference denotes clean-view MRR with hard candidates. (B) Clean-view minus leave-one-modality-out MRR with hard candidates. MS/MS comprises both polarity channels. Supporting values for panels A and B are reported in Tables S10 and S6, respectively.}
\label{fig:6}
\end{figure}

To interpret individual modality contributions, we considered two complementary analyses for MoE-mixed under hard candidates. First, complete-input MRR was 0.9867, exceeding every single-modality result (Fig.~\ref{fig:6}A). The single-modality order was \textsuperscript{13}C~NMR (0.9647), \textsuperscript{1}H~NMR (0.9626), IR (0.9307), and MS/MS (0.8575). Second, leave-one-modality-out evaluation showed that removing any modality reduced MRR when the remaining evidence was retained (Fig.~\ref{fig:6}B): the decreases were 0.0054 for \textsuperscript{13}C~NMR, 0.0050 for \textsuperscript{1}H~NMR, 0.0032 for IR, and 0.0005 for MS/MS, calculated before rounding. Thus, Fig.~\ref{fig:6}A measures how well each modality supports ranking alone, whereas Fig.~\ref{fig:6}B measures its additional contribution alongside the other spectra. Their joint pattern is consistent with complementary and partly redundant evidence: MS/MS supported substantial discrimination alone, with a smaller incremental contribution when IR and both NMR modalities were already available. These effects characterize the fitted model and candidate set~\cite{Geurts2016improving}.

However, the apparent value of individual modalities also depended on the structures being compared. Replacing hard candidates with random candidates increased MRR for all four single-modality inputs while leaving the MoE-mixed checkpoint unchanged (Fig.~\ref{fig:6}A and Table~S10). The largest increase occurred for MS/MS, from 0.8575 to 0.9377, with a paired gain of +0.0802. The \textsuperscript{1}H~NMR score rose from 0.9626 to 0.9874, a gain of +0.0248, while gains for \textsuperscript{13}C~NMR and IR were +0.0248 and +0.0242, respectively. Although the modality order was retained, the gap between MS/MS and the other modalities narrowed. NMR-only scores under random candidates slightly exceeded the clean-view hard-candidate reference, illustrating the importance of candidate composition when comparing input configurations.

The accompanying composition statistics further characterize this change in task difficulty (Table~S11). In the hard-negative set, the median absolute exact-mass difference from the correct structure was $2.6854\times10^{-3}$~Da, compared with 101.0867~Da for random negatives. In addition, the proportion sharing the formula of the correct structure fell from 14.2538\% to 0.0007\%, and the proportion sharing a non-empty Murcko scaffold decreased from 4.2770\% to 0.7943\%. These changes occurred together with a reduction in median fingerprint similarity. Given that fragmentation patterns can constrain molecular substructures and frameworks, the larger MS/MS gain is compatible with reduced ambiguity among the random alternatives~\cite{stein2012libraries}. Mass, formula, scaffold, and similarity changed jointly. Therefore, the comparison quantifies the overall effect of candidate composition.

The above results highlight the joint role of input conditions and candidate difficulty in study design and support two conclusions. First, complete multimodal evidence improved ranking within hard sets that prioritized chemically related alternatives, whereas random candidates made each single-modality task easier. Second, evaluation with hard candidates exposed differences in discrimination that were less apparent under broad random sampling. Accordingly, any modality ordering should be interpreted together with its model, training strategy, and candidate construction. The strict first-rank comparisons in Section~S3.4 provide additional context for this interpretation.

\subsection{Illustrative reranking cases}

\begin{figure}[H]
    \centering
    \includegraphics[width=\linewidth,height=0.68\textheight,keepaspectratio]{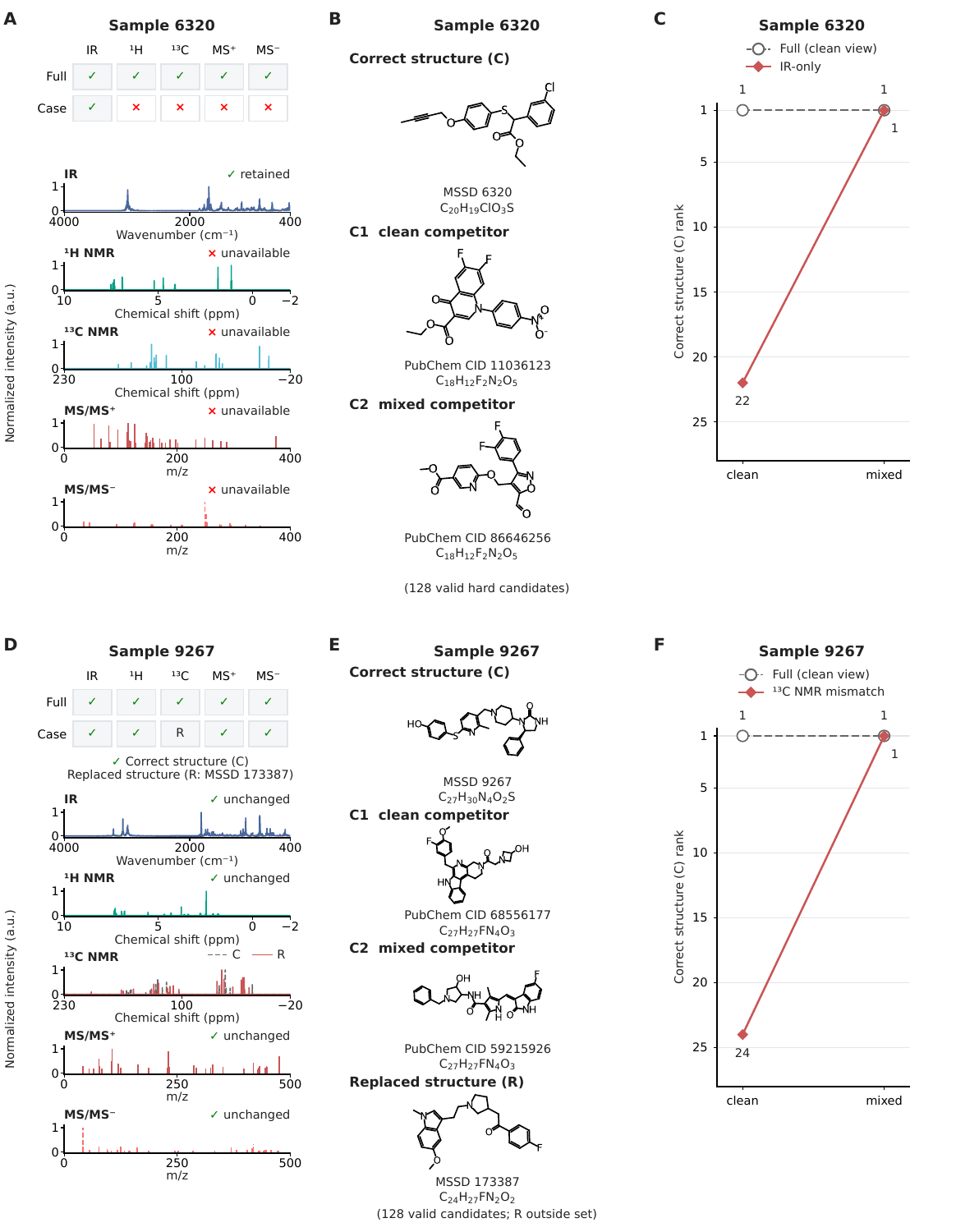}
    \caption{Post hoc reranking cases: (A--C) IR-only input for sample~6320; (D--F) \textsuperscript{13}C~NMR replacement for sample~9267. Columns show spectra, structures, and correct-structure ranks. C denotes the correct structure; C1/C2 denote the top-scoring negatives for MoE-clean/MoE-mixed under the altered input; R denotes the replacement source, outside the candidate set. Unavailable spectra and the original \textsuperscript{13}C~NMR trace are references. Both models use identical evidence and 128 fixed candidates per case.}
    \label{fig:7}
\end{figure}

To illustrate how the population-level gains translated into individual reranking decisions, two post hoc cases were examined under sparse and inconsistent spectral evidence, as shown in Fig.~\ref{fig:7}.

Firstly, an IR-only input case was examined (Fig.~\ref{fig:7}A--C). For MSSD sample~6320, both MoE models ranked the correct structure (C) first when all five channels were available. By contrast, with only the IR spectrum retained, MoE-clean placed the correct structure at rank~22, whereas MoE-mixed retained it as the unique first-ranked candidate among 128 valid structures. The correct structure had formula $\mathrm{C}_{20}\mathrm{H}_{19}\mathrm{Cl}\mathrm{O}_{3}\mathrm{S}$ and contained an ester, an alkynyl ether substituent, and an aryl thioether linkage. The highest-scoring MoE-clean competitor, PubChem compound identifier (CID)~11036123, had formula $\mathrm{C}_{18}\mathrm{H}_{12}\mathrm{F}_{2}\mathrm{N}_{2}\mathrm{O}_{5}$ and a different heteroaromatic framework. Under mixed training, this competitor moved to rank~29. Thus, the different bonding environments in Fig.~\ref{fig:7}B provide a chemically plausible basis for the IR-based discrimination shown in Fig.~\ref{fig:7}A and the rank change in Fig.~\ref{fig:7}C. The shift therefore involved discrimination between structurally distinct, mass-matched alternatives using IR evidence alone, while the candidate identities and spectrum of the correct structure were held fixed.

Secondly, an inconsistent molecular-source case was examined (Fig.~\ref{fig:7}D--F). For MSSD sample~9267, the correct structure (C) had formula $\mathrm{C}_{27}\mathrm{H}_{30}\mathrm{N}_{4}\mathrm{O}_{2}\mathrm{S}$ and was ranked first by both models under clean input. The conflict view replaced only its \textsuperscript{13}C~NMR spectrum with that of MSSD molecule~173387, selected based on similar functional groups. The replacement source (R) had formula $\mathrm{C}_{24}\mathrm{H}_{27}\mathrm{F}\mathrm{N}_{2}\mathrm{O}_{2}$, differing from that of the correct structure despite being selected based on functional-group similarity. The overlaid traces in Fig.~\ref{fig:7}D document the altered carbon-environment evidence, while Fig.~\ref{fig:7}E establishes that R was outside the candidate set. With the other four channels from the correct structure retained, MoE-clean placed the correct structure at rank~24 and PubChem CID~68556177 first. By contrast, MoE-mixed ranked the correct structure uniquely first and that competitor at rank~21. The 128 valid candidates were identical across both models and both input views, and none of the correct-structure ranks involved tied scores. Thus, Fig.~\ref{fig:7}F illustrates correct-candidate recovery under a fixed source mismatch.

Taken together, these two cases illustrate how mixed-condition training supports candidate discrimination under substantially incomplete spectral input and preserves the prioritization of the correct structure when a single spectral channel is inconsistent with the remaining evidence.

\section{Conclusion}
\label{sec:conclusion}

A chemistry- and spectroscopy-informed mixed-condition training strategy was developed to improve the robustness of multimodal candidate structure reranking. In a controlled two-by-two factorial comparison using spectra from 79,462 held-out samples evaluated across 30 predefined views, the training strategy was identified as the principal source of performance improvement across both fusion architectures. Relative to clean training, mixed training increased the 30-view mean MRR by 6.08\% for MoE and 5.69\% for Concat. For the MoE architecture, mixed training also increased R@1 by 6.86 percentage points (7.67\% relative). The largest improvements occurred under incomplete or mismatched inputs, while performance with complete, unaltered inputs remained high. Mixed training also improved calibration and sample-level mismatch detection. The modality-specific robustness patterns were further discussed in the context of their underlying chemical and spectroscopic characteristics, offering a domain-informed interpretation of the different responses of MS/MS, IR, and NMR to spectral perturbations. Moreover, the comparison between hard and random candidate sets demonstrated that single-modality reranking performance depends strongly on candidate-set composition, underscoring the importance of realistic candidate-space evaluation. Overall, the proposed training strategy and fixed-candidate evaluation framework provide a reproducible chemometric approach to more reliable molecular structure prioritization from complementary spectra, conditional on inclusion of the correct structure in the candidate set. For future studies, further development and validation using large-scale datasets of experimentally measured multimodal spectra, together with improved localization of inconsistent channels, will be important for advancing this approach toward practical application.

\section*{Code and data availability}

The Multimodal Spectroscopic Dataset (MSSD v3) used in this study is publicly available at \url{https://doi.org/10.5281/zenodo.14770232}.

The code will be made publicly available after publication of the article in a peer-reviewed journal.

\section*{CRediT author statement}

Bowen Gao: Conceptualization, Methodology, Software, Validation, Formal analysis, Investigation, Data curation, Visualization, Writing -- original draft, Writing -- review and editing.

Lei Zhu: Funding acquisition, Resources, Supervision, Writing -- review and editing.

Yiying Wang: Resources, Supervision, Project administration, Writing -- review and editing.

Wenjie Yu: Resources, Supervision, Project administration, Writing -- review and editing.

\section*{Generative AI disclosure statement}

During the preparation of this manuscript, the authors used ChatGPT (OpenAI) for language editing and grammar refinement to improve readability. All AI-generated content was reviewed and edited by the authors. The authors hold full responsibility for all content of the publication.

\section*{Acknowledgments}

This study was supported by the AI for Science Program, the Shanghai Municipal Commission of Economy and Informatization (2025-GZL-RGZN-BTBX-02027). The funder played no role in study design, data collection, analysis and interpretation of data, or the writing of this manuscript.

\bibliographystyle{elsarticle-num}
\bibliography{references}

\end{document}